\documentclass[10pt, a4paper]{article}
\usepackage{ltc05}
\usepackage{todonotes}

\usepackage{subcaption}
\usepackage{array}
\newcommand*{\YePersian}{\includegraphics[scale=0.047]{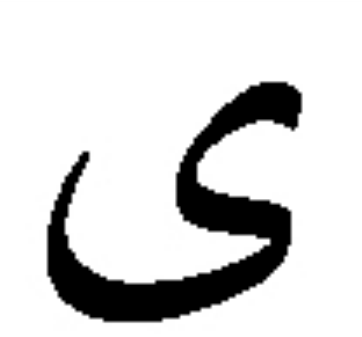}}

\newcommand*{\naawandakaani}{\includegraphics[scale=0.7,trim =0mm 1mm 0mm 0mm]{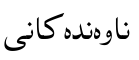}}
\newcommand*{\naawand}{\includegraphics[scale=0.7,trim =0mm 1mm 0mm 0mm]{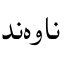}}
\newcommand*{\aan}{\includegraphics[scale=0.7,trim =0mm 1mm 0mm 0mm]{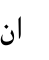}}

\newcommand*{\dangdaanmaan}{\includegraphics[scale=0.7,trim =0mm 1mm 0mm 0mm]{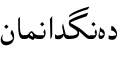}}
\newcommand*{\dangdaan}{\includegraphics[scale=0.7,trim =0mm 1mm 0mm 0mm]{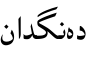}}
\newcommand*{\maan}{\includegraphics[scale=0.7,trim =0mm 1mm 0mm 0mm]{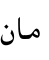}}

\newcommand*{\krdn}{\includegraphics[scale=0.7,trim =0mm 1mm 0mm 0mm]{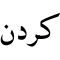}}
\newcommand*{\dastgir}{\includegraphics[scale=0.7,trim =0mm 1mm 0mm 0mm]{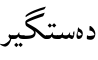}}
\newcommand*{\asir}{\includegraphics[scale=0.7,trim =0mm 1mm 0mm 0mm]{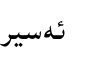}}

\newcommand*{\aka}{\includegraphics[scale=0.7,trim =0mm 1mm 0mm 0mm]{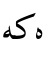}}

\newcommand*{\noosi}{\includegraphics[scale=0.7,trim =0mm 1mm 0mm 0mm]{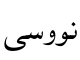}}
\newcommand*{\noos}{\includegraphics[scale=0.7,trim =0mm 1mm 0mm 0mm]{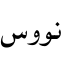}}
\newcommand*{\noosin}{\includegraphics[scale=0.7,trim =0mm 1mm 0mm 0mm]{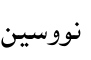}}
\newcommand*{\n}{\includegraphics[scale=0.01]{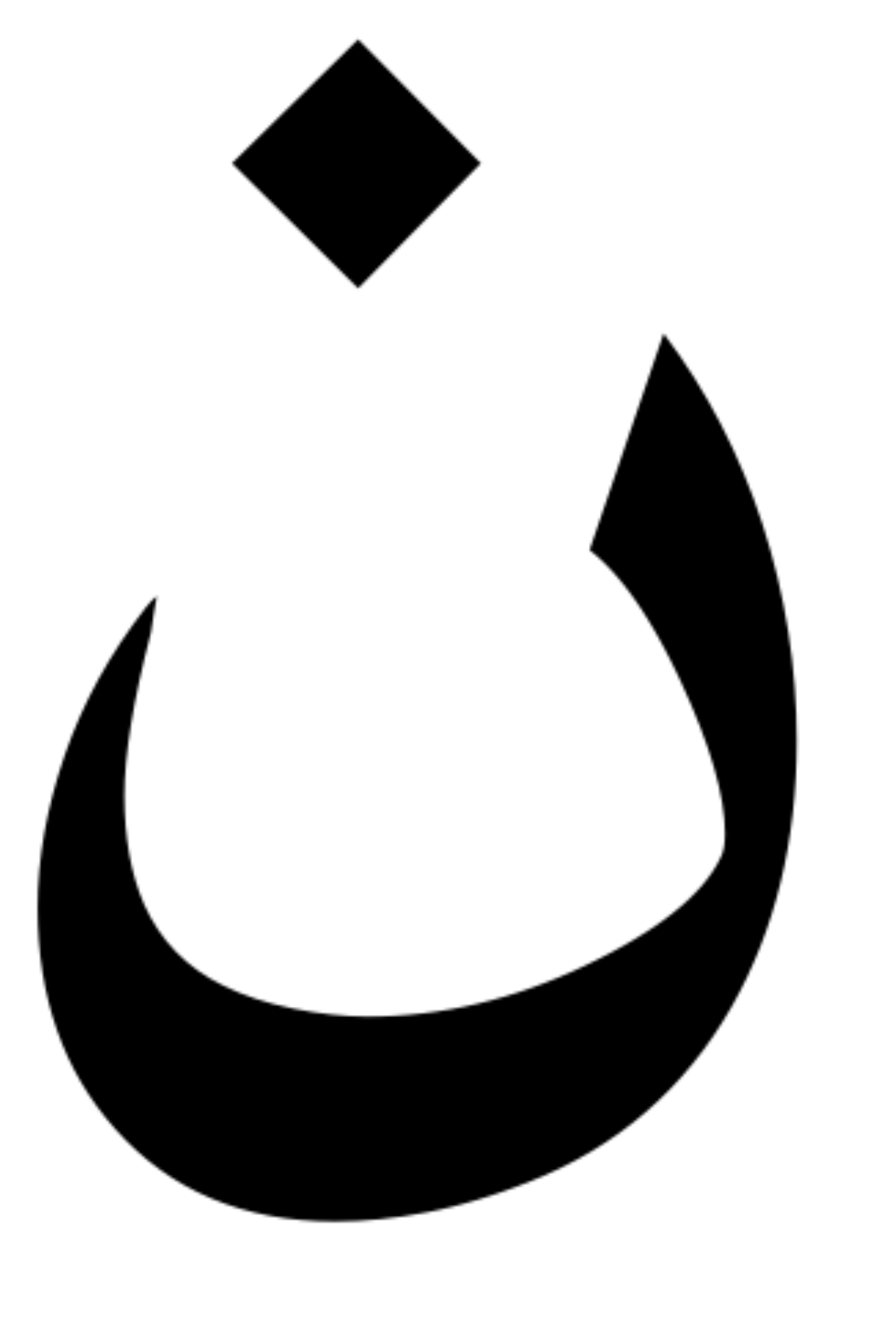}}

\newcommand*{\naa}{\includegraphics[scale=0.01]{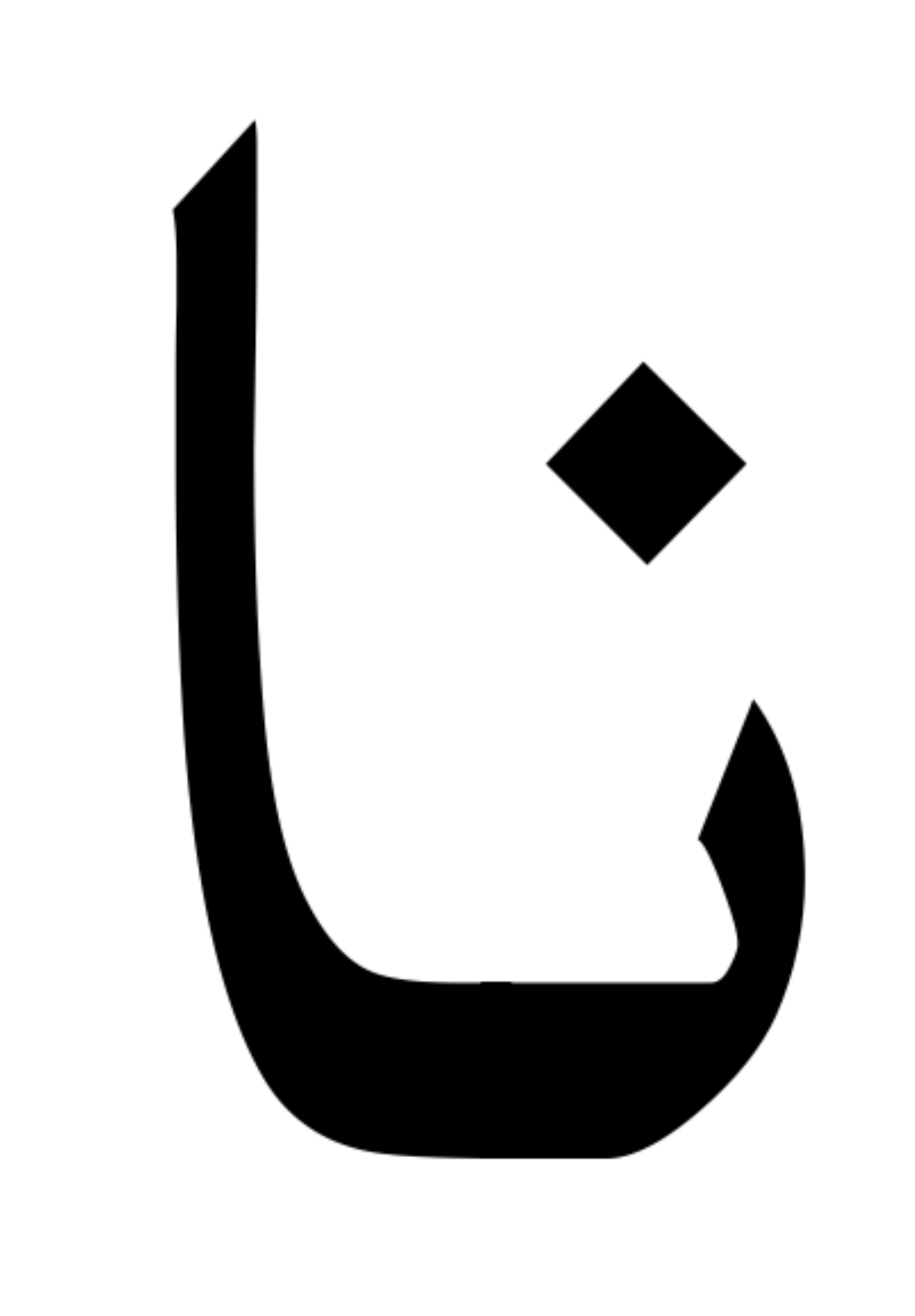}}
\newcommand*{\na}{\includegraphics[scale=0.01]{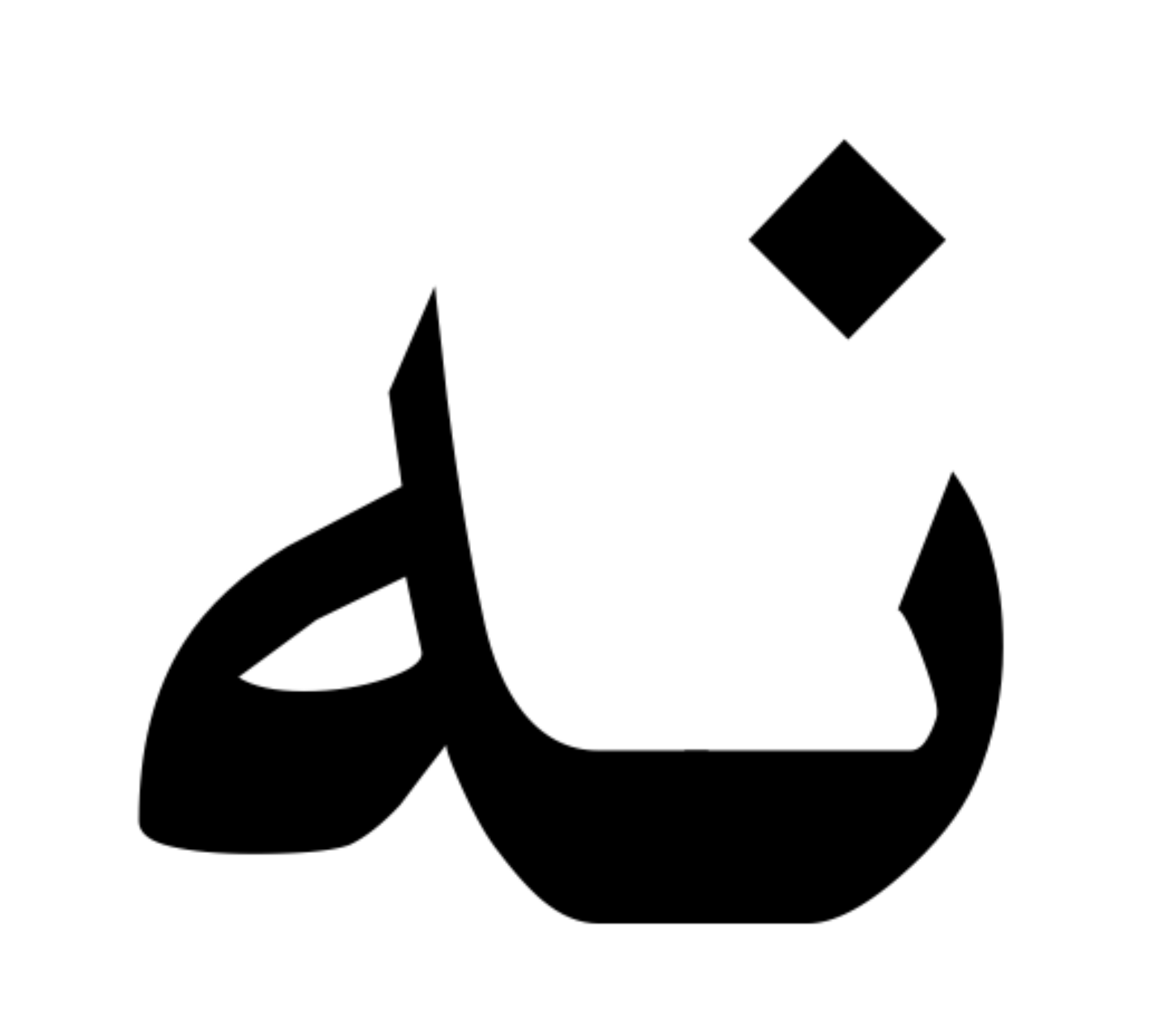}}

\newcommand*{\halwest}{\includegraphics[scale=0.7,trim =0mm 1mm 0mm 0mm]{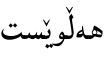}}
\newcommand*{\otomobel}{\includegraphics[scale=0.7,trim =0mm 1mm 0mm 0mm]{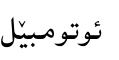}}

\newcommand*{\lawlat}{\includegraphics[scale=0.7,trim =0mm 1mm 0mm 0mm]{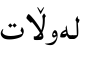}}
\newcommand*{\lawlati}{\includegraphics[scale=0.7,trim =0mm 1mm 0mm 0mm]{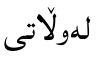}}
\newcommand*{\la}{\includegraphics[scale=0.7,trim =0mm 1mm 2mm 0mm]{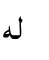}}

\newcommand*{\nishtan}{\includegraphics[scale=0.7,trim =0mm 1mm 1mm 0mm]{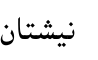}}
\newcommand*{\nisht}{\includegraphics[scale=0.7,trim =0mm 1mm 0mm 0mm]{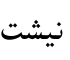}}

\newcommand*{\nishtman}{\includegraphics[scale=0.7,trim =0mm 1mm 0mm 0mm]{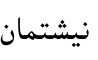}}

\usepackage[english]{babel} 
\usepackage{multirow}
\usepackage[utf8]{inputenc}
\usepackage{hyperref}
\usepackage{wrapfig}
\usepackage{enumitem}
\usepackage{algorithm}
\usepackage[noend]{algpseudocode}
\usepackage{amsmath}
\usepackage{tikz}
\usetikzlibrary{shapes}
\usetikzlibrary{arrows}
\usetikzlibrary{positioning}
\usepackage{booktabs} 
\usepackage{graphicx}
\usepackage{pgfplots}
\pgfplotsset{compat=1.10}
\usepackage{wrapfig}
\usepackage{multirow}
\usepackage{chngcntr}

\title{Building a Lemmatizer and a Spell-checker for Sorani Kurdish}
\name{Shahin Salavati$^{\ast}$, Sina Ahmadi$^{\dagger}$} 
\address{ $^{\ast}$University of Kurdistan \\
               Sanandaj, Iran \\ 
               shahin.salavati@ieee.org \\ \\
               $^{\dagger}$Paris Descartes University \\ 
               45, rue des Saints-Pères, 75006 Paris \\
               sina.ahmadi@etu.parisdescartes.fr}

\abstract{
The present paper aims at presenting a lemmatization and a word-level error correction system for Sorani Kurdish. We propose a hybrid approach based on the morphological rules and a n-gram language model. We have called our lemmatization and error correction systems \textit{Peyv} and \textit{Rênûs} respectively, which are the first tools presented for Sorani Kurdish to the best of our knowledge. The Peyv lemmatizer has shown 86.7\% accuracy. As for Rênûs, using a lexicon, we have obtained 96.4\% accuracy while without a lexicon, the correction system has 87\% accuracy. As two fundamental text processing tools, these tools can pave the way for further researches on more natural language processing applications for Sorani Kurdish.
}

\begin{document}

\maketitleabstract

\section{Introduction}\label{sec:intro}

Due to the increasing proliferation of digital texts in nowadays life, the field of text processing has caught the attention of various researchers. Despite extensive efforts in some most spoken languages like English, the body of researches done in the field of text processing in Kurdish is still scanty. Kurdish is called a \textit{dialect continuum}, meaning that it contains a series of different dialects, among which the two major dialects are Kurmanji and Sorani\cite{KurdishMacKenzie}. The focus of this paper is on the Sorani dialect.

Following our previous work on a stemming system, introduced as \textit{Jedar} for Sorani Kurdish \cite{salavati2013stemming,KLPPStemmer}, in the current study we present \textit{Peyv} as a lemmatization system.  The goal of both stemming and lemmatization is to reduce different forms of words in order to obtain a common base form. In the stemming systems, this common base form is called stem and is extracted by reducing derivationally related forms of a word. In the other hand, lemmatization concerns the inflectional forms of a given word to reduce to a base or dictionary form of the word, i.e. lemma.  

Morphological analysis, applying lexicon and reconstruction after removal of affixes are among the methods common to lemmatization algorithms. Considering the possible prefixes and suffixes attaching to a lemma, Sorani has a complex structure in terms of morphology. As a result, Sorani Kurdish lemmatization algorithms are to have a far more accurate mechanism. Our proposed lemmatization algorithm puts to use the Sorani Kurdish morphological rules in order to extract lemmas of words. These rules have been extracted for different parts of speech such as noun, verb and adjective. 

One of the most common applications of lemmatizers must be sought in spell-checker algorithms. In a spell checking system, for a given potentially wrong word, the system proposes a list of possible corrections in a ranking form. In this paper we present \textit{Rênûs}, as the first Sorani Kurdish spell checker.

Our proposed systems are based on a n-gram language model generated from Pewan \cite{KLPPCorpus} text corpus containing 18M words from 115K news articles\cite{esmaili2013sorani}. We have also used a rule-based method based on the morphological rules of Sorani Kurdish. These two methods have been used for different languages so far. In a recent study, n-gram language model has been used for CoNLL 2014 Shared Task for Grammatical Error Correction\cite{hernandez2014conll}.

The rest of the paper is organized as follows. First, we give a brief description of the Sorani Kurdish, particularly its morphology in section \ref{sec:background}. Then in sections \ref{sec:payv} and \ref{sec:renoos}, we explain our methods to construct the Peyv lemmatizer and the Rênûs spell checker respectively. The results of our experiments are reported in section \ref{sec:experiments} and the paper is concluded in section \ref{sec:conclusion}. Two test set examples may be found in appendix \ref{appendix} as well. In order to provide a more convenient reading for those who are less familiar with the Arabic-based orthography of Sorani Kurdish, the transliteration of each meaningful word in the Latin-based orthography along with the translation is provided in parentheses. 

\section{Background}\label{sec:background}

As a Western Iranian language, Kurdish shares the same characteristic of having a very limited amount of synthetic verbal lexemes (around 300) \cite{walther2012fitting}. Morphologically speaking, Kurdish verbs are formed by adding different inflectional and derivational set of prefixes and suffixes clustered around a given stem, so that a variety of affixes in some cases even gets to procreate more than 100 words derived from one stem.

Various structures of words are generated by two concepts of word formation, that is inflectional and derivational processes. Inflectional process adapts the word to match its grammatical category in the sentence. The declension in gender, number, tense and person is an inflectional process. Forming the word "flowers" out of "flower", is an inflectional change implying the number. Usually these changes do not alter the root in any fundamental way. In contrast, they adapt and homogenize the sentences in terms of grammar. For example, in the sentence "flowers are happy things", the plural form of "flower" and the verb "are" match together. These inflectional morphemes are productive, in a way that they can be applied to a great number of words.

The second concept is the derivational process that often involves the addition of a derivational suffix or other affix which applies to words of one lexical category and forms new words of another such category. It is differentiated from inflectional process by changing the meaning of the root without changing grammatical structure of the sentence. For example, "potter" and "pottery" are two different words with different meanings formed by derivational process. Therefore, in the sentence “I saw the potter/pottery", there is no change in the other parts of the sentence. In most cases in Sorani Kurdish, derivational affixes are positioned after the root and before the inflectional affixes, like "potteries".

	\begin{figure}[h]
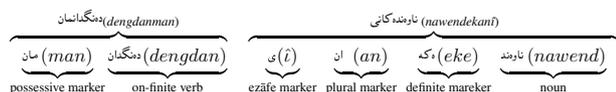

	\centering
	\scalebox{.7}{
	$\overbrace{  \underbrace{\maan (man)}_\textrm{possessive marker} \underbrace{	\dangdaan (dengdan)}_\textrm{on-finite verb}}^{\text{\dangdaanmaan (\textit{dengdanman})}}\thinspace\thinspace\thinspace\thinspace\thinspace\thinspace \overbrace{\underbrace{\YePersian (\hat{\iota})}_\textrm{ezāfe marker} \thinspace\thinspace \underbrace{\aan (an)}_\textrm{plural marker} \thinspace\thinspace \underbrace{\aka (eke)}_\textrm{definite mareker} \thinspace\thinspace \underbrace{\naawand (nawend)}_\textrm{noun}\thinspace\thinspace\thinspace\thinspace }^{\text{\naawandakaani (\textit{nawendekanî})}}$
}
	\caption{Lexical items in an example nominal group \textit{"nawendekanî dengdanman"} (our voting centers)}
	\label{figure:1}
	\end{figure}
	
Figure \ref{figure:1} represents components of the phrase \{ \dangdaanmaan \naawandakaani \} (\textit{nawendekanî dengdanman}, "our voting centers") which is consisted of a noun \{\naawand \} (\textit{nawend} "center") with three suffixes, and a non-finite verb \{\dangdaan \} (\textit{dengdan} "voting") with a suffix. Table \ref{table:2} provides some of the frequent nominal affixes of Sorani Kurdish as well. Table \ref{table:3} also shows some frequent verbal affixes in Sorani dialect. Different possibilities for endings are seperated by "-". Note that there is no affix for gender in Sorani.

\begin{table}[h]
\centering
  \includegraphics[width=\linewidth]{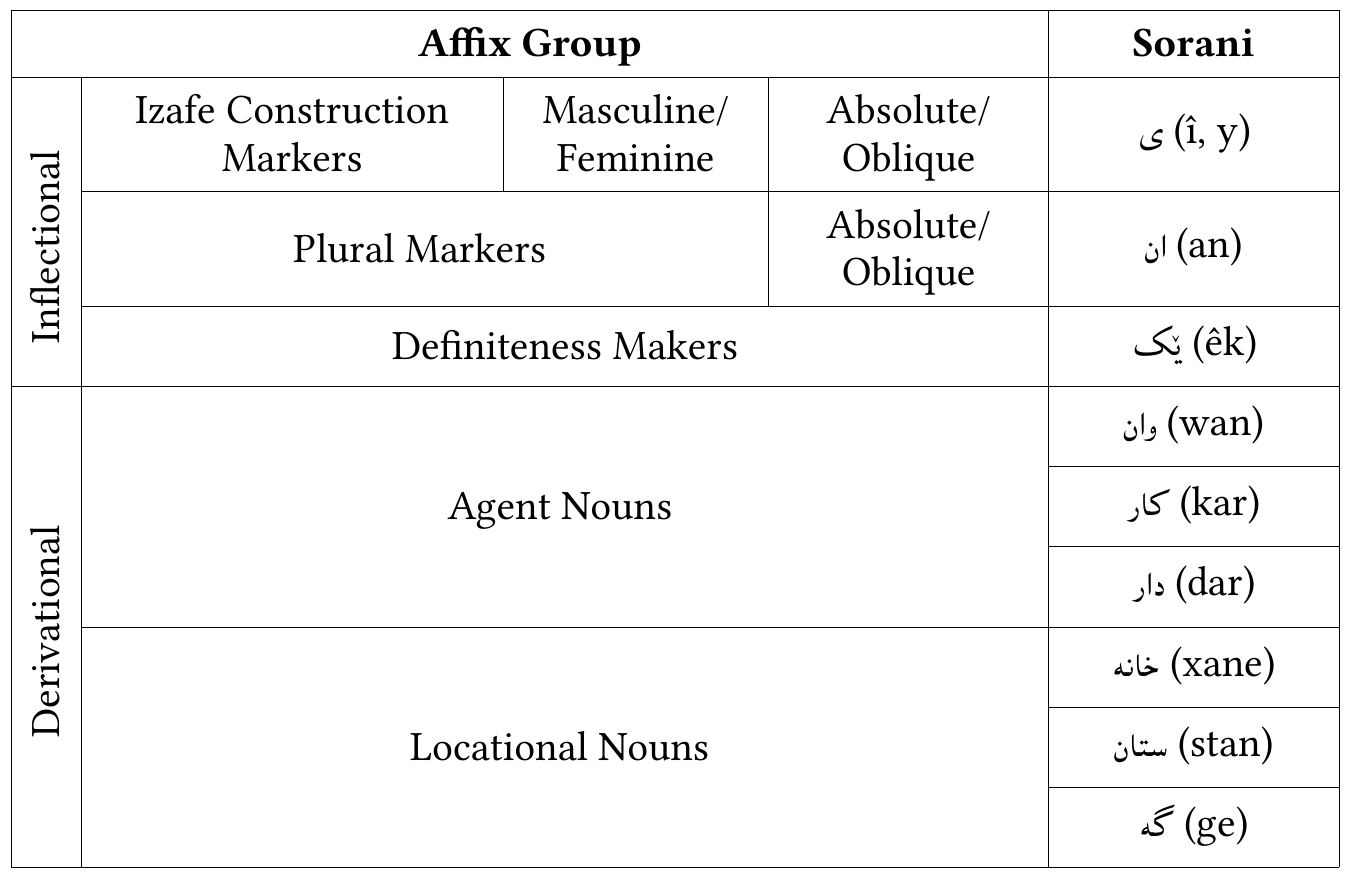}
  \caption{Frequent nominal affixes in Sorani Kurdish}
    \label{table:2}
\end{table}

\begin{table}[h]
    \centering
      \includegraphics[width=\linewidth]{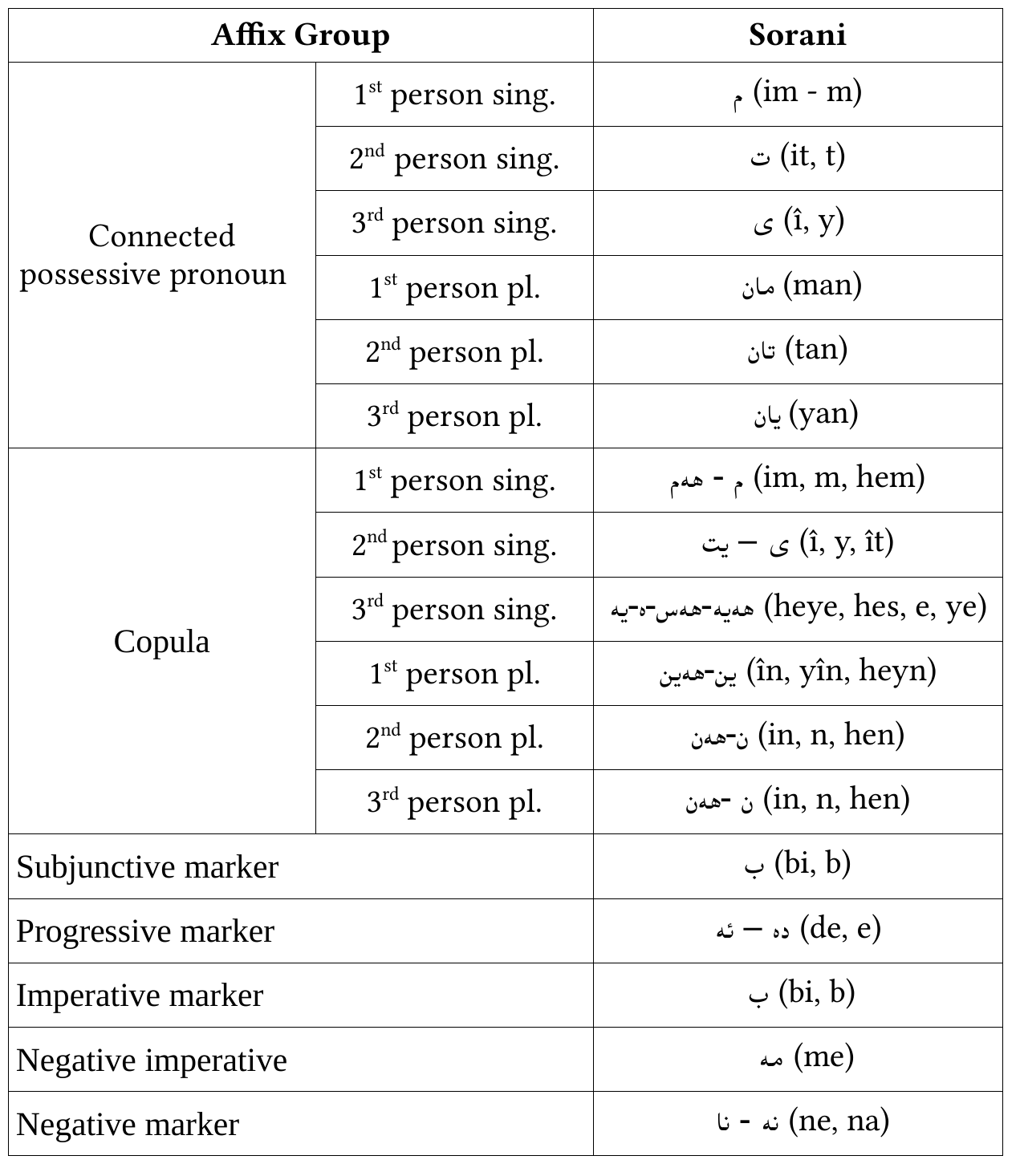}
	\caption{Frequent verbal affixes in Sorani Kurdish}
	\label{table:3}
\end{table}
    
Auxiliary verbs have an important role in Sorani Kurdish in order to form new verbs belonging to different semantic categories. For example, the verb \{\krdn\}(\textit{kirdin} "to do") provides different meanings combining to other nouns, e.g., \{\krdn\dastgir\} (\textit{"destgîr kirdin"} to arrest), \{\krdn\asir\} (\textit{"esîr kirdin"} to capture). 

The conjugation of Kurdish verbs due to the different inflections based on tense, person, positivity or negativity, transitivity or in-transitivity can be a complex task with different exceptions. The inflections are brought about by adding different prefixes and suffixes to the root word of the verb. Therefore, in order to get the root of a given verb, a more accurate procedure is required. 

There are two tenses for Sorani Kurdish verbs: past and present (non-past). The past verbs have five different types while the present ones have only two and the imperative and negative imperative inflections are the major inflections of active voice. The significant point is that there are two distinct roots for these two tenses, meaning that different inflections derive from different roots of verbs. For example, the infinitive \{\noosin\} (\textit{nûsîn} to write) has a past stem that is \{\noosi\} (\textit{nûsî}) and a present one that is \{\noos\} (\textit{nûs}). The past stem is derived from the current infinitive regularly by omitting the final letter \{\n\} (\textit{n/in}) which is the case for all Kurdish verbs; however, to find the present root there is no specific pattern. Since there is no future tense in Sorani Kurdish, adverbs are generally used in present tense to express future tense.

\begin{table*}[htbp]

	\centering
		\includegraphics[scale=0.6]{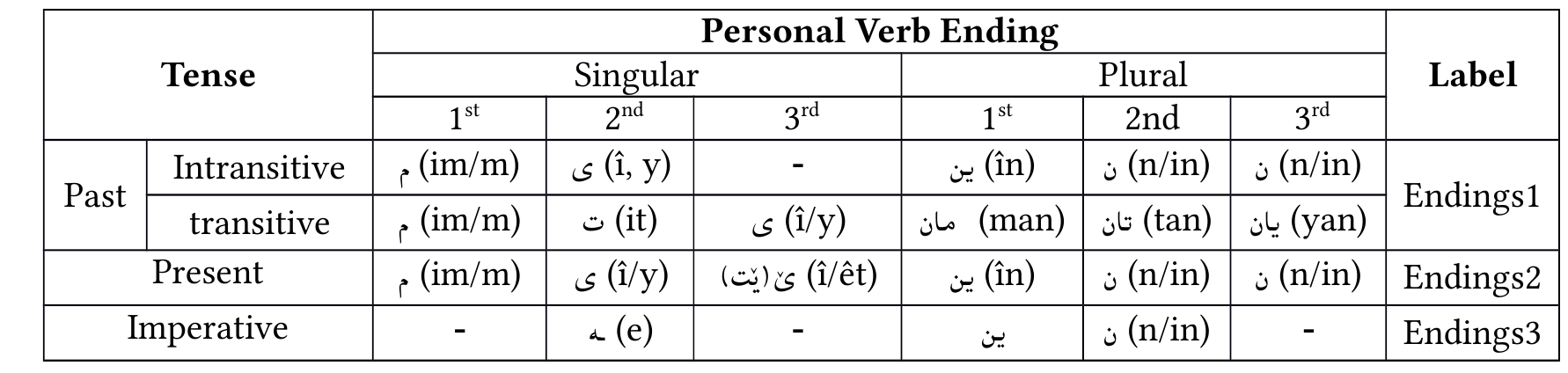}
	\caption{The most frequent verb endings in Sorani Kurdish}
	\label{table:4}
	
\end{table*}

Another important point about verbs is "person". There are 6 persons in the Sorani, three singular and three plural. The persons in either one of the tenses have their own particular endings which could be observed in table \ref{table:4}. There are two categories of verb endings for the past tense depending on the transitivity of the verb. Transitive verbs endings are the same enclitics and similar to possessive adjectives. Note that these endings represent a major part of Sorani Kurdish verbs and not all. 

In an attempt to formalize Sorani Kurdish verbs structure, we have provided table \ref{table:5} for the most frequent cases. It should also be noted that the negative form of verbs in Sorani is made by adding \{\naa\} (\textit{na}) or \{\na\} (\textit{ne}) to the verb. Some of the most frequent verbal prefixes of Sorani are as follows: \{ \includegraphics[scale=0.75,trim =0mm 4mm 0mm 0mm]{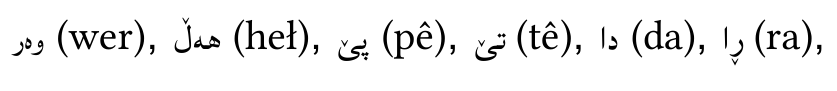} \\  \includegraphics[scale=0.75,trim =0mm 4mm 0mm 0mm]{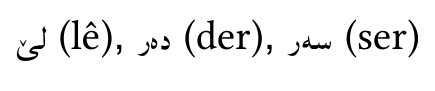} \}


\begin{table}[htbp]
	\centering
		\scalebox{.9}{
		\includegraphics[scale=0.6]{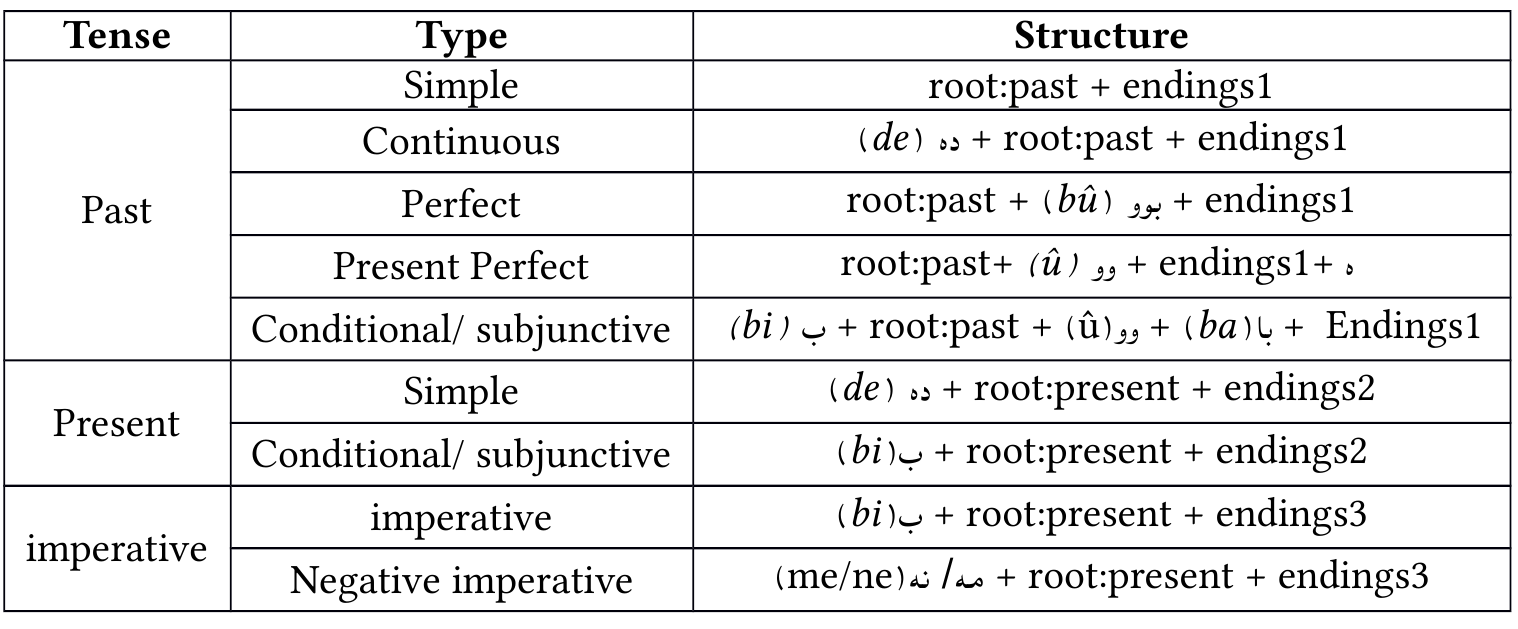}
		}
	\caption{A formalization of the most common rules in Sorani Kurdish verbs structure}
	 \label{table:5}
\end{table}

\section{The Peyv Lemmatizer}\label{sec:payv}

After extracting the basic morphological rules in Sorani Kurdish, we have implemented the Peyv lemmatizer for nouns and verbs respectively. In the case of nouns, a pruning method is used to find the root. In the other hand, a bottom-up method is suggested for the verbs which have more complex structures than nouns, e.g., multiple inflective morphemes without order. Lexical items have been detected using a tagged lexicon for verbs and nouns.

\subsection{Noun lemmatization}

Considering the possibility of attaching multiple affixes to Kurdish words, Peyv lemmatizes each noun in a recursive way by omitting the affixes. In order to distinguish the affixes, a list of 45 common inflectional affixes has been collected. The words of less than three characters in length have been put out of the algorithm. In addition, in order to guarantee the performance of the lemmatizer, 1500 words, mostly exceptions or non-Kurdish words, have been defined. These words end in subcategories that match the inflectional affixes. Note that the derivational affixes are not omitted in this algorithm.

\begin{figure}[b]

	\centering
		\includegraphics[scale=0.3]{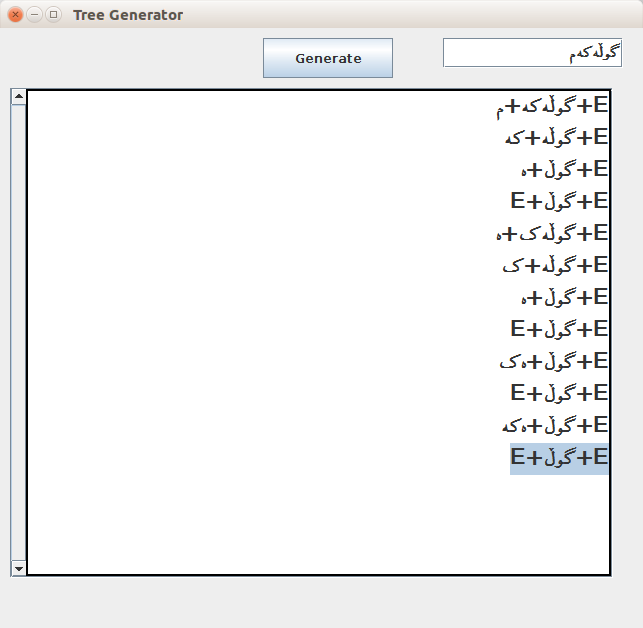}
	\caption{Peyv lemmatizer has detected the root of the input noun "gułekem" as "guł" after pruning the inflectional morphemes. "E" refers to empty morphemes. }
	\label{fig:pruned_tree}
	
\end{figure}

Figure \ref{fig:pruned_tree} shows the output of the Peyv lemmatizer for a given noun \{\includegraphics[scale=0.7,trim =0mm 1mm 0mm 0mm]{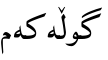}\} (\textit{gułekem} my flower). The system has detected all the possible affixes in an arborescent form, and finally has suggested the possible root, \{\includegraphics[scale=0.7,trim =0mm 1mm 0mm 0mm]{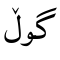}\} (\textit{guł} flower), based on the predefined morphological rules. "E" has been used to represent empty suffixes or prefixes.


%
%

\subsection{Verb lemmatization}

What makes the lemmatization of the verbs different from that of the nouns is the more complex orderless structure. Attaching of prefixes and suffixes to the root of the verbs and their probable change of meaning complicates even more the process of lemmatization. Therefore, the common process of omitting the suffix is not efficient.

As a rule-based method, we have provided a list of the past and present roots of Sorani Kurdish. The method used in this algorithm is in the increasing or bottom-up form, meaning that the likely roots for the given word are identified in the list of the existing roots based on the congruence of the subcategories. Then, different markers and prefixes of the root are detected based on the rules. If the created verbs accords with the entered word, the current root will be reverted as the head of the verb. Peyv has put to analysis the past, present, imperative, negative imperative, passive verbs and other verbal structures such as compound verbs and auxiliary verbs.

\section{The Rênûs Spellchecker}\label{sec:renoos}

Spell-checking correction is generally done in two stages: error detection and error correction. Error detection includes methods to lookup a language model to detect an error. In the other hand, the task of error-correction is to generate the most likely correct word-forms given a misspelled word-form. We have split this task in two different tasks: generating suggestions and ranking them.

In the first step, we have created our language model based on the grams and their related frequencies in Pewan corpus. The basic idea is production of the n-grams of the input word and calculation of the gram frequency in the data set. A word with lowest frequency among the n-grams, i.e., $frequency=0$, would be detected as a wrong word. A rule-based method would also be used to look up a lexicon. Table \ref{table:7} shows the difference between frequencies of two words \{\includegraphics[scale=0.7,trim =0mm 1mm 0mm 0mm]{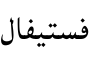}\} and \{\includegraphics[scale=0.7,trim =0mm 1mm 0mm 0mm]{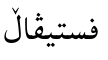}\} (\textit{fistîvał} festival). The correct form of the word is marked in bold. 

\begin{table}[t]

	\centering
	\scalebox{.9}{
		\includegraphics[scale=0.6]{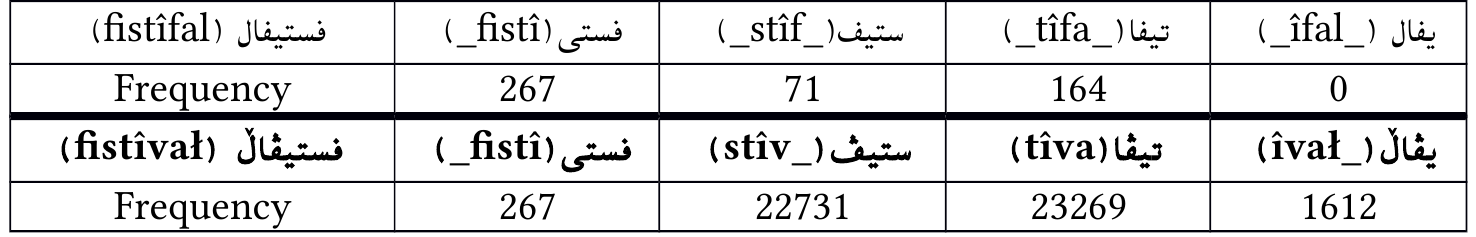}}
	\caption{Error detection based on the n-gram frequency for an example word, festival}
	\label{table:7}
	
\end{table}

After detecting a potentially wrong word, its grams are created and the lowest frequency gram will be chosen. The given word would be corrected based on the edit distance, as a measure of similarity, and frequency. Finally, a candidate gram from the list of suggestions is proposed as the corrected form of the word. We have provided some meta information for the correction system, including orthographic preprocessing for Arabic-based orthography of Sorani. In the case that a given word include a mistakenly spelled character, we reduce the cost value in the edit distance to $distance = 0.5$ instead of $distance=1$. Characters have been classified based on this distance value in 6 groups as follows: \{ \includegraphics[scale=0.7,trim =0mm 2.5mm 0mm 0mm]{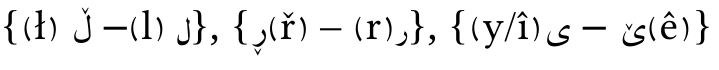} \\ \includegraphics[scale=0.72,trim =0mm 2.5mm 0mm 0mm]{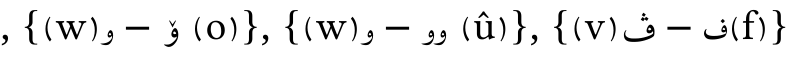} \}

After generating corrections, all the grams with an edit distance of less than 3  would be inserted in the candidate list. Then, we rank the candidates based on the edit distance and frequency of each one, shown in equation  \ref{eq1fgfg}. 


\begin{equation}\label{eq1fgfg}
\resizebox{1\hsize}{!}{$%
 Rank(\widehat{T}) = \left\{
  \begin{array}{l l}
    \frac{1}{dist(\widehat{T}, T)}\times freqValue(\widehat{T}) & \quad if (dist(\widehat{T}, T) \leq 1) \\
   \frac{1}{(dist(\widehat{T}, T)-1)\times \alpha}\times freqValue(\widehat{T}) & Otherwise
  \end{array} \right.
  $%
}%
\end{equation}

where $dist(T, \widehat{T})$ represents the edit distance between the candidate word $\widehat{T}$ and the input word $T$, and $freqValue(\widehat{T})$ represents the frequency of the suggested gram for the candidate word. $\alpha$ is a variable to bias the placement of the words with high standard frequency and distance at the top of the list. Particularly, since the nominal affixes have a high frequency, if not considering this parameter, they are more likely to be suggested as the correction candidates.

Equation \ref{eq1fgfg} also indicates that similarity is inversely correlated with distance and directly with frequency. This equation has given priority to the edits distances less than 1. The main idea of using a standard frequency is to rate the common words. The maximum effect of this parameter according to the equation is on the candidates that have the same standard frequency. In other words, among candidates with equal edit distance, the priority goes with the ones with a higher standard frequency.

Another factor that makes this algorithm more efficient is the position of the grams in the word which are classified in three groups: beginning, middle and end. Thus, to suggest a candidate, the search will be conducted only among the grams of the same position. It should be noted that this priority has been followed in the production of grams and frequency of collection of documents.

In order to provide more trustworthy implementation, we have ignored the words with low frequency, i.e., 1 or 2. Pewan corpus contains collected online documents that are not essentially gold-standard. Therefore, to create the grams, these set of words were omitted. Evidently, within this process some correct words will also be removed, although this does not affect the grams in a remarkable way, since they could be produced out of other words. 
\section{Experiments}\label{sec:experiments}

\subsection{Peyv lemmatization}

In this section, we present results of the proposed methods. In order to evaluate the efficiency of Peyv lemmatizer, a set of derivatives of two random words \{\otomobel\} (\textit{utumibêl} automobile) \ref{table:automobile} and \{\halwest\} (\textit{he\l{}wîst} decision) \ref{table:halwist} have been collected using Pewan corpus. The algorithm detected wrong roots in 6 cases (out of 120) for  \{\halwest\} derivatives, and 4 cases (out of 38) for \{\otomobel\}. Thus, the algorithm has an accuracy of 95\% and 89.4\% for lemmatization of these two words, respectively. Having said that the wrongly lemmatized roots have not been correctly spelled in the source word.

We have evaluated Peyv on verb test sets as well. The lemmatizer finds the correct root of a verb in the case that the root exists in the lexicon. In this set of verbs, different structures consisting past and present tense, imperative and negative imperative, passive and negative are included. The exceptions also have been taken into account. 

\begin{table}[h]
\centering
\scalebox{0.75}{
\begin{tabular}{|l|l|l|l||l|}
\hline
              & True correction & False correction & Total & Accuracy \\ \hline
$1^{st}$ group     & 173            & 30              & 203   & 0.8522   \\ \hline
$2^{nd}$ group     & 289            & 40              & 329   & 0.878    \\ \hline
$3^{rd}$ group     & 420            & 69              & 489   & 0.860    \\ \hline
Pewan queries & 122            & 14              & 136   & 0.901    \\ \hline \hline
Total         & 1004           & 153             & 1157  & 0.867    \\ \hline
\end{tabular}
}
	\caption{Peyv results. Different test sets have been used to insure randomness of data.}
	\label{table:peyvresults}
\end{table}

Table \ref{table:peyvresults} shows the results of lemmatization, after being double-checked manually. In order to guarantee randomness of the test set, we have evaluated the Peyv lemmatizer in three of words from Pewan corpus. Also, the set of unique vocabulary of Pewan corpus queries has been tested. The average accuracy of the word sets is 86.7\%. In most of the cases, writing errors have been the reason of detection of a wrong root for a given word. For instance, Peyv returns \{\lawlat\} (\textit{lewi\l{}at}) as the root of a wrong spelled word, \{\lawlati\} (\textit{lewi\l{}atî}), while the correct root should be detected without the prefix \{\la\} (\textit{le} from). A very large quantity of Kurdish lemmas would be needed in order to provided an ideal lemmatizer in working on the exceptions. The accuracy of the system is calculated by dividing the true predictions by the test set size.

\subsection{Rênûs spell checker}

Evaluation of the spell checker is based on the comparison of the first-ranked predictions and the the gold-standard words. Given a set of $N$ words, $G_i$ is the gold-standard correction of word $i$ and $P_i$ is the first-ranked prediction of the correction system for word $i$. We define accuracy as follows:

\begin{equation}
accuracy = \frac{ | G \cap P|}{|G|} 
\end{equation}

where the intersection between $G$ and $P$ indicates number of matches between the gold-standard corrections and the first-ranked predictions of the system. Since our error correction system is word-level, the gold-standard correction set $G$ and the first-ranked predictions $P$ would definitely have the same size, thus, $|G|=|P|$.

We have evaluated the correction system with a lexicon and also without using a lexicon as follows.

\begin{table}[]
\centering
\scalebox{0.8}{
\begin{tabular}{|l|l|l|l|l|l|l|}
\hline
Method   & \multicolumn{2}{l|}{raw grams} & \multicolumn{2}{l|}{gram/frequency} & \multicolumn{2}{l|}{most frequent words} \\ \hline
3-gram   & 292           & 18\%           & 767              & 48\%             & 799                      & 50\%          \\ \hline
4-gram   & 691           & 43\%           & 1188             & 74\%             & 1233                     & 77\%          \\ \hline
5-gram   & 963           & 60\%           & 1254             & 79\%             & 1305                     & 82\%          \\ \hline
4,5-gram & 996           & 62\%           & 1334             & 84\%             & \textbf{1392}          & \textbf{87\%}          \\ \hline\hline
average  & 753           & 46\%           & 1135             & 71\%             & 1182                     & 74\%          \\ \hline
\end{tabular}
}
\caption{Rênûs spell checker results without using lexicon}
	\label{table:12}
\end{table}

\subsubsection{Error correction without lexicon}

Table \ref{table:12} shows the results of Rênûs spell checker without using a lexicon. These results are based on n-grams method. The second column, raw grams, refers to the use of the whole extracted n-grams of Pewan corpus without refining. The column gram/frequency refers to the extracted grams with a frequency higher than 2. Finally, the last column is the set of the most frequent words and the three groups mentioned in table \ref{table:peyvresults}. These last two columns show that the removal of low-frequency words has improved the results of the system. The 4,5-gram as the combination of 4-grams and 5-grams has finally shown the highest accuracy of 87\%.

\subsubsection{Error correction with lexicon}

We have also evaluated the usage of lexicon along with the n-grams language model for the task of error correction. The results presented in table \ref{table:13} represent the evaluation results based on the same classification used in \ref{table:12}. The system presents the most accurate results in the 4,5-gram test set where the accuracy gets to 96.4\%.

In several cases, the wrong lemmatization of a word led to a wrong correction. For instance, the wrong word \{\nishtan\} which is suggested as the correct form of \{\nishtman\} (\textit{nîştiman} homeland), is wrongly lemmatized by the Peyv lemmatizer, suggesting \{\nisht\} (nîşt) as its root.

\begin{table}[h]
\centering
\scalebox{0.8}{
\begin{tabular}{|l|l|l|l|l|l|l|}
\hline
Method   & \multicolumn{2}{l|}{raw grams} & \multicolumn{2}{l|}{gram/frequency} & \multicolumn{2}{l|}{most frequent words} \\ \hline
3-gram   & 1417           & 89\%          & 1459             & 92\%             & 1463                    & 92\%           \\ \hline
4-gram   & 1446           & 91\%          & 1505             & 94\%             & 1512                    & 95\%           \\ \hline
5-gram   & 1473           & 92\%          & 1513             & 95\%             & 1517                    & 95.7\%         \\ \hline
4,5-gram & 1475           & 93\%          & 1523             & 96\%             & \textbf{1529}         & \textbf{96.4\%}         \\ \hline\hline
average  & 1452           & 91\%          & 1500             & 94\%             & 1182                    & 94\%           \\ \hline
\end{tabular}
}
\caption{Rênûs spell checker results using lexicon}
\label{table:13}
\end{table}

%
%
%

Back to equation \ref{eq1fgfg}, we have also evaluated the impact of different values for $\alpha$. $\alpha$ is a variable to bias the placement of the words with high standard frequency and edit distance. In the case that $\alpha=1$, frequency is directly applied with no impact of $\alpha$. Table \ref{table:15} shows the evaluation results in regards to this parameter. Candidates with bigger edit distance and higher frequency are more probable to be the correct form of a given word.

\begin{table}[h]
\centering
\scalebox{0.7}{
\begin{tabular}{|l|l|l|l|l|l|}
\hline
$\alpha$                    & Number of predictions & 3-gram & 4-gram & 5-gram & 4,5-gram \\ \hline
\multirow{4}{*}{1}   & 1              & 0.20   & 0.33   & 0.42   & 0.42     \\ \cline{2-6} 
                     & 2              & 0.288  & 0.43   & 0.54   & 0.55     \\ \cline{2-6} 
                     & 5              & 0.41   & 0.57   & 0.68   & 0.69     \\ \cline{2-6} 
                     & 10             & 0.509  & 0.68   & 0.75   & 0.78     \\ \hline
\multirow{4}{*}{10}  & 1              & 0.27   & 0.43   & 0.52   & 0.53     \\ \cline{2-6} 
                     & 2              & 0.389  & 0.56   & 0.65   & 0.67     \\ \cline{2-6} 
                     & 5              & 0.53   & 0.68   & 0.75   & 0.78     \\ \cline{2-6} 
                     & 10             & 0.63   & 0.78   & 0.80   & 0.84     \\ \hline
\multirow{4}{*}{70}  & 1              & 0.28   & 0.47   & 0.47   & 0.56     \\ \cline{2-6} 
                     & 2              & 0.40   & 0.59   & 0.59   & 0.698    \\ \cline{2-6} 
                     & 5              & 0.55   & 0.71   & 0.71   & 0.80     \\ \cline{2-6} 
                     & 10             & 0.66   & 0.80   & 0.81   & 0.854    \\ \hline
\multirow{4}{*}{100} & 1              & 0.28   & 0.47   & 0.55   & 0.57     \\ \cline{2-6} 
                     & 2              & 0.40   & 0.59   & 0.67   & 0.69     \\ \cline{2-6} 
                     & 5              & 0.55   & 0.71   & 0.76   & 0.78     \\ \cline{2-6} 
                     & 10             & 0.66   & 0.79   & 0.80   & 0.852    \\ \hline
\multirow{4}{*}{200} & 1              & 0.28   & 0.47   & 0.55   & 0.57     \\ \cline{2-6} 
                     & 2              & 0.40   & 0.59   & 0.66   & 0.69     \\ \cline{2-6} 
                     & 5              & 0.55   & 0.71   & 0.76   & 0.80     \\ \cline{2-6} 
                     & 10             & 0.65   & 0.79   & 0.80   & \textbf{0.86}     \\ \hline
\end{tabular}
}
	\caption{Accuracy of the Rênûs spell checking system in regards to different values of $\alpha$ (see equation \ref{eq1fgfg})}
	\label{table:15}
\end{table}

\section{Conclusion and Future Work}\label{sec:conclusion}

In  this  paper,  we  introduced two fundamental language processing tools for Sorani Kurdish: Peyv lemmatizer and Rênûs spell-checker. These tools are the first existing tools for Sorani. We have used a hybrid approach based on the extracted morphological rules and a n-gram language model. In the n-gram method, we have analyzed 18M words from 115K news articles from Pewan text corpus. We have also provided a tagged lexicon for detecting lexical items (nouns and verbs). 

The Peyv lemmatization has shown 86.7\% accuracy. In most of the cases, writing errors have been the reason of a wrong lemmatization. Our experimental  results  also indicate the role of lemmatization which can greatly improve the quality of Rênûs spell checker. Using a lexicon, we have obtained 96.4\% accuracy, while without a lexicon, Rênûs spell checker has 87\% accuracy. We have also observed that the removal of low-frequency words improves the results of the system.

The implemented tools have been inevitable parts of most of the researches led by KLPP\footnote{Kurdish Language Processing Project: \url{http://eng.uok.ac.ir/esmaili/research/klpp/en/main.htm}}. These tools may pave the way for further researches on Kurdish as a less-resourced language.

\nocite{*}
\bibliographystyle{ltc05}
\bibliography{refs} 

\newpage
\appendix
\section{APPENDIX}
\label{appendix}

\begin{table}[h]
	\centering
	\scalebox{0.8}{
		\includegraphics[scale=0.6]{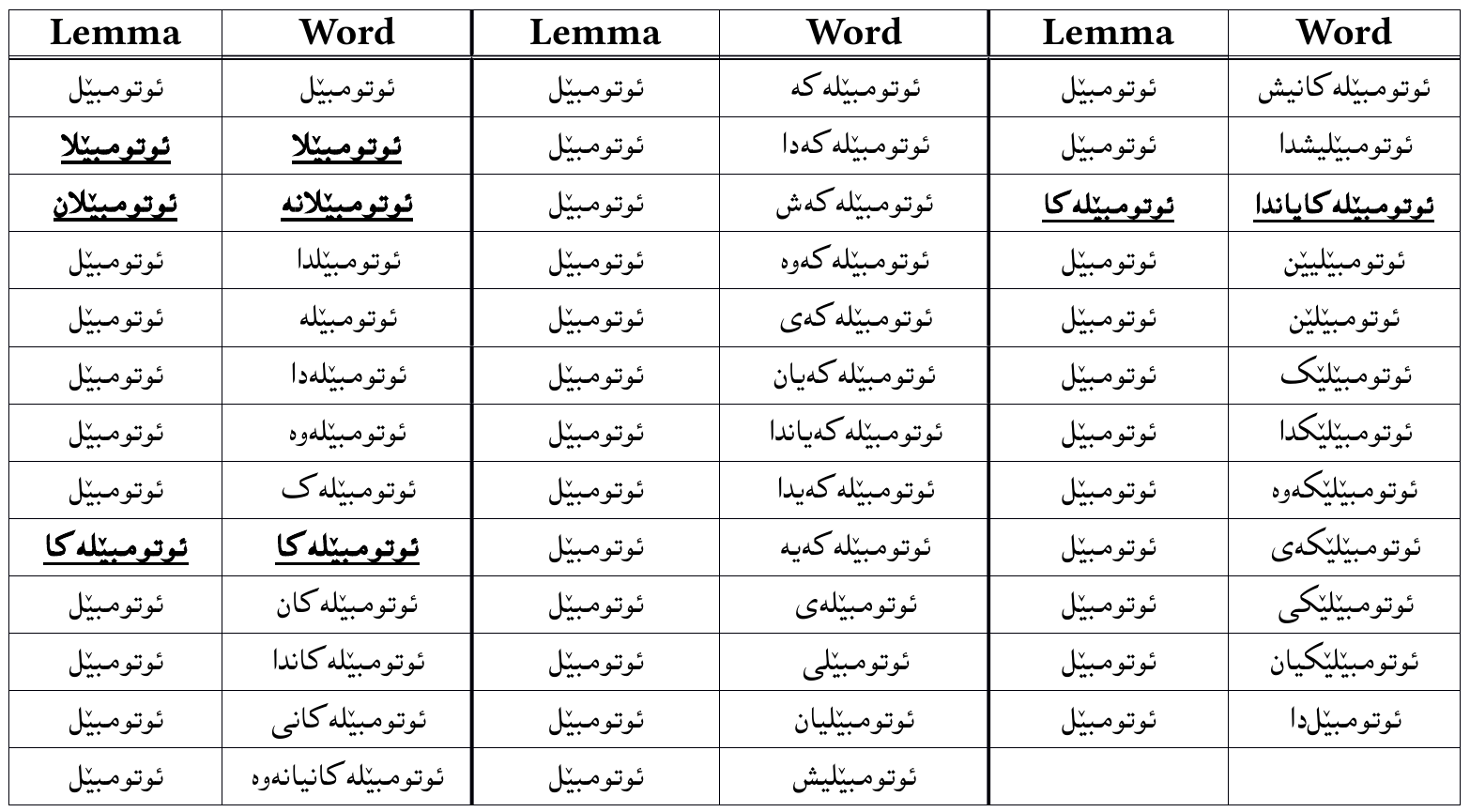}
		}
	\caption{ Lemmatization of \textit{utumibêl}. 4 cases among all are lemmatized wrongly. }
	\label{table:automobile}
	
\end{table}

\begin{table}[h]
	\centering
	\scalebox{0.8}{
		\includegraphics[scale=0.5]{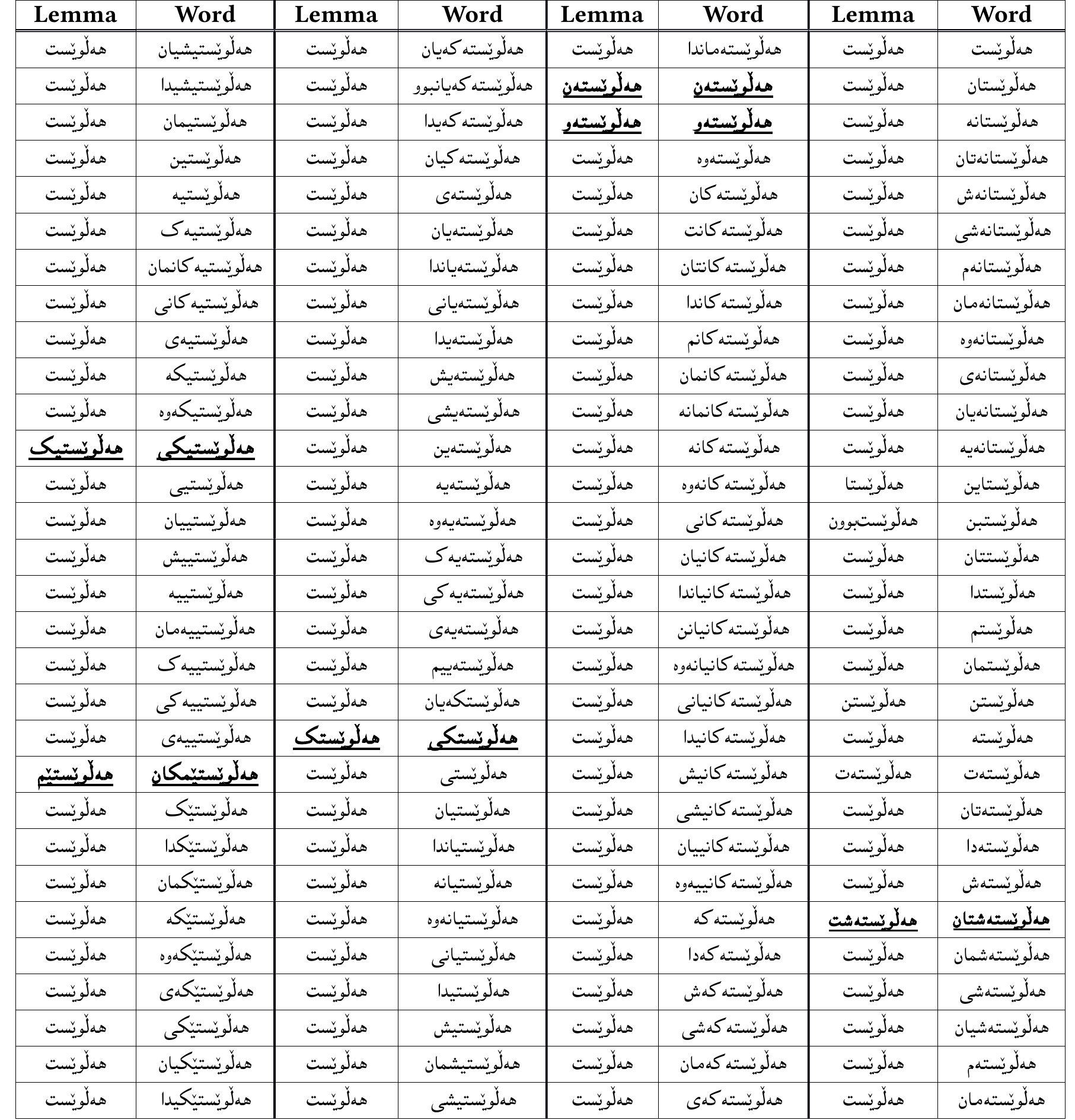}
		}
	\caption{ Lemmatization of \textit{he\l{}wîst}. 6 wrongly lemmatized words have been specified in bold.}
	\label{table:halwist}
	
\end{table}

\end{document}